\documentclass[10pt,twocolumn,letterpaper]{article}

\usepackage[pagenumbers]{cvpr} 

\usepackage{booktabs}
\usepackage{multirow}
\usepackage{makecell}
\usepackage{xcolor} 
\usepackage{amssymb}
\usepackage{caption}
\usepackage{tabularx}
\usepackage{booktabs}  
\usepackage{siunitx}   
\usepackage{multirow}  
\usepackage{algorithm}
\usepackage{algpseudocode}
\usepackage{amsmath} 

\definecolor{darkgreen}{RGB}{0,128,0}
\definecolor{darkred}{RGB}{178,34,34}

\usepackage{pifont}
\newcommand{\cmark}{\ding{51}} 
\newcommand{\xmark}{\ding{55}} 

\renewcommand{\paragraph}[1]{\vspace{1pt}\noindent\textbf{#1}}

\definecolor{dark_green}{rgb}{0, 0.5, 0}
\definecolor{red}{rgb}{1.0, 0, 0}

\definecolor{cvprblue}{rgb}{0.21,0.49,0.74}
\usepackage[pagebackref,breaklinks,colorlinks,allcolors=cvprblue]{hyperref}

\def\paperID{584} 
\def\confName{CVPR}
\def\confYear{2026}

\title{VideoMaMa: Mask-Guided Video Matting via Generative Prior}

\author{
    Sangbeom Lim\textsuperscript{1$^\dagger$} \qquad
    Seoung Wug Oh\textsuperscript{2} \qquad
    Jiahui Huang\textsuperscript{2} \qquad
    Heeji Yoon\textsuperscript{3} \\[0.5em]
    Seungryong Kim\textsuperscript{3} \qquad
    Joon-Young Lee\textsuperscript{2} \\[0.8em]
    \textsuperscript{1}Korea University \qquad
    \textsuperscript{2}Adobe Research \qquad
    \textsuperscript{3}KAIST AI\\[0.8em]
    {\tt \href{https://cvlab-kaist.github.io/VideoMaMa}{\textcolor{cvprblue}{https://cvlab-kaist.github.io/VideoMaMa}}}
}

\begin{document}
\twocolumn[{%
\renewcommand\twocolumn[1][]{#1}%
\maketitle
\centering
    \vspace{-15pt}
    \includegraphics[width=1\linewidth]{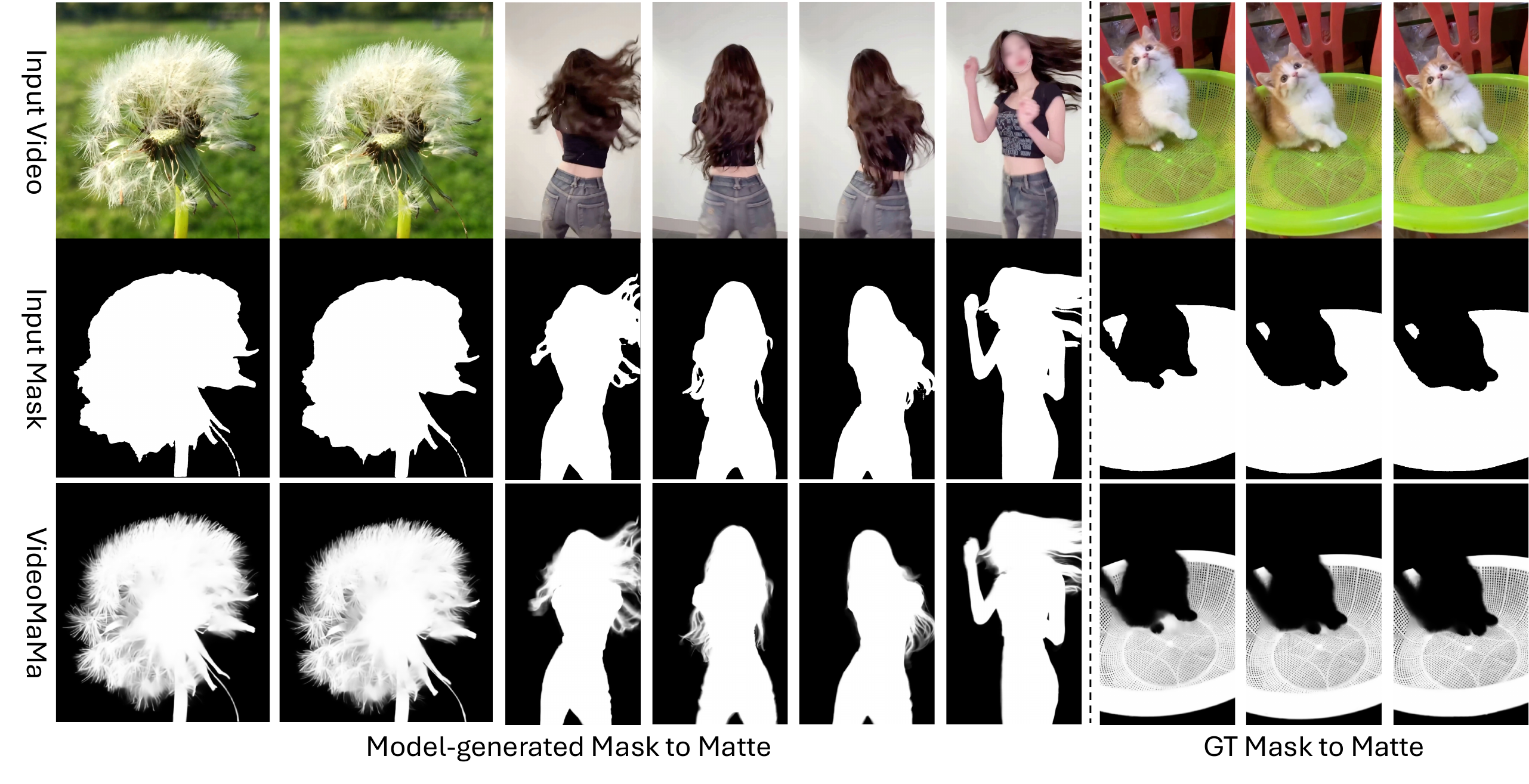}
    \vspace{-20pt}
    \captionof{figure}{We introduce Video Mask-to-Matte Model (\textbf{VideoMaMa}), a diffusion-based model that generates high-quality alpha mattes from input binary segmentation masks obtained either from existing models such as SAM2~\cite{sam2} or from ground-truth segmentation masks in existing datasets such as SA-V~\cite{sam2}. 
    Examples shown highlights our VideoMaMa's ability to capture fine-grained details including motion blur, and intricate boundary structures on natural video footage.
    }
\label{fig:teaser}
\vspace{1em}
}
]

\renewcommand{\thefootnote}{\fnsymbol{footnote}} 
\footnotetext[2]{Work done during an internship at Adobe Research.} 
\renewcommand{\thefootnote}{\arabic{footnote}} 

\begin{abstract}
Generalizing video matting models to real-world videos remains a significant challenge due to the scarcity of labeled data. 
To address this, we present Video Mask-to-Matte Model (\textbf{VideoMaMa}) that converts coarse segmentation masks into pixel accurate alpha mattes, by leveraging pretrained video diffusion models. VideoMaMa demonstrates strong zero-shot generalization to real-world footage, even though it is trained solely on synthetic data. 
Building on this capability, we develop a scalable pseudo-labeling pipeline for large-scale video matting and construct the Matting Anything in Video (\textbf{\mbox{MA-V}}) dataset, which offers high-quality matting annotations for more than 50K real-world videos spanning diverse scenes and motions. 
To validate the effectiveness of this dataset, we fine-tune the SAM2 model on \mbox{MA-V} to obtain SAM2-Matte, which outperforms the same model trained on existing matting datasets in terms of robustness on in-the-wild videos.
These findings emphasize the importance of large-scale pseudo-labeled video matting and showcase how generative priors and accessible segmentation cues can drive scalable progress in video matting research.
\end{abstract}    
\section{Introduction}
\label{sec:intro}

Video matting, the task of extracting foreground objects with pixel-level precision from video, serves as a fundamental component in video editing applications, including background replacement~\cite{zeng2025lumen, gao2025anyportal, ren2024relightful}, visual composition~\cite{dalva2024layerfusion, yang2025generative}, and relighting~\cite{chaturvedi2025synthlight, ren2024relightful, wang2025comprehensive}. 

Despite its importance, training a video matting model that performs robustly across diverse real-world videos remains challenging for two key reasons.
First, high-quality video matting annotations are extremely scarce. Ground-truth mattes are typically captured in controlled environments such as chroma key studios~\cite{video240k, video108} or with specialized camera setups~\cite{enomoto2025polarized}, which makes it difficult to scale annotation efforts and to capture diverse objects, scenes, and camera angles. As a result, existing datasets are often limited to human portraits.

Second, most video matting models are trained on synthetic videos, where foregrounds are composited onto arbitrary backgrounds. Such synthetic compositions often introduce unrealistic artifacts in lighting, motion blur, and temporal coherence.
These two factors, limited data diversity and the domain gap between synthetic and real videos, significantly hinder the ability of current models to generalize to real-world footage where complex interactions between foregrounds and backgrounds naturally occur.

In this work, we propose a novel bootstrapping strategy that effectively bridges the gap between synthetic and real-world videos, enabling more robust video matting.
We introduce Video Mask-to-Matte Model (\textbf{VideoMaMa}), a diffusion-based model that converts binary segmentation masks into continuous alpha mattes (see Figure~\ref{fig:teaser}). 
Built upon pretrained video diffusion models~\cite{svd}, VideoMaMa demonstrates remarkable zero-shot generalization to real-world videos, even though the model is trained exclusively on synthetic data.
This strong generalization ability stems from diffusion priors trained on internet-scale image and video data, which can generate high-quality content across diverse domains~\cite{ltxvideo, wan, svd}. 
To adapt these priors for video matting while preserving their generative capabilities, we introduce a two-stage training strategy that separately optimizes spatial and temporal layers, along with semantic knowledge injection via DINOv3~\cite{dinov3} features.

To evaluate the robustness of the VideoMaMa model, we conduct experiments on video matting guided by binary masks from diverse sources, including those produced by existing video segmentation models.
The results show that VideoMaMa consistently generates high-quality video matting outputs regardless of the input mask type, demonstrating strong robustness.
These findings suggest that VideoMaMa can be effectively utilized as a pseudo-labeler for constructing large-scale video matting annotations.

In addition, by leveraging VideoMaMa, we propose a simple but scalable pipeline for constructing large-scale video matting annotations and introduce the Matting Anything in Videos (\textbf{\mbox{MA-V}}) dataset, the first large-scale pseudo video matting dataset built by converting segmentation labels from the SA-V dataset~\cite{sam2}. 
Unlike existing video matting datasets~\cite{maggie, GVM, matanyone, video108, video240k}, which rely on synthetic or composited content, \mbox{MA-V} provides high-quality matting annotations for over 50K real-world videos covering diverse scenes, objects, and motion dynamics, as summarized in Table~\ref{tab:dataset_stats}. 
This scalable pipeline demonstrates the potential of leveraging generative priors and segmentation labels, which are considerably easier to obtain than video alpha matte, to efficiently construct large-scale, high-quality video matting datasets.

To validate the effectiveness of large-scale pseudo video matting annotations, we fine-tune the SAM2~\cite{sam2} model for video matting using \mbox{MA-V}, referred to as SAM2-Matte.
Trained without any architectural modifications, the SAM2-Matte model achieves substantially more robust matting performance on in-the-wild videos than the same SAM2 model fine-tuned on existing video matting datasets, as well as other existing video matting methods.
The results demonstrate the strong potential of large-scale pseudo annotations to drive advancements in video matting research.

Our contributions can be summarized as follows.
\begin{itemize}
    \item \textbf{VideoMaMa}: A diffusion-based model that generates realistic video matting annotations from binary masks, enabling scalable matting label creation with easily obtainable segmentation labels.
    \item \textbf{\mbox{MA-V} Dataset}: The first large-scale, high-quality video matting dataset built on real captured footage, consisting of over 50K diverse videos.
\end{itemize}
\section{Related Work}
\label{sec:relwork}

\paragraph{Video matting.}
Video matting methods can be categorized into auxiliary-free and auxiliary-guided approaches. Auxiliary-free methods~\cite{lin2022robust, modnet, vmformer} typically focus on portrait matting, restricting their applicability to human-centric scenarios. Trimap-guided approaches~\cite{huang2023end, otvm} require manual trimap annotations, limiting their practicality for zero-shot inference. Recent mask-based guidance methods address these limitations: MaGGIe~\cite{maggie} decouples tracking from matting using binary mask tracks, MatAnyone~\cite{matanyone} propagates target-assigned matting through a memory-augmented architecture, and GVM~\cite{GVM} employs diffusion models for portrait matting. Despite these advances, existing methods remain constrained by domain specificity.

Beyond algorithmic limitations, existing datasets also constrain progress. Representative datasets include VideoMatte240K~\cite{video240k} (484 videos), VideoMatting108~\cite{video108} (108 videos), and VM800~\cite{matanyone} (826 videos), predominantly focusing on human subjects in controlled settings~\cite{DVM, crgnn, GVM}. Critically, all rely on composition-based generation where foreground objects are composited onto random backgrounds, creating artificial scene compositions. No large-scale real video matting dataset exists where foreground and background naturally co-occur. 

\paragraph{Diffusion models for perception tasks.}
Recent advances have demonstrated that diffusion models~\cite{stablediffusion, svd, wan, hunyuanvideo} encode rich priors about natural scenes, motion dynamics, and temporal coherence. 
Marigold~\cite{marigold} pioneered this direction by finetuning Stable Diffusion for monocular depth estimation, achieving strong zero shot generalization despite training only on synthetic data. This success has extended to semantic segmentation~\cite{diception, mmgen} and video depth estimation~\cite{depthcrafter, chronoDepth}.
These models exhibit remarkable zero shot generalization across perception tasks~\cite{depthcrafter, diception, efe, genpercept, lotus}, maintaining strong performance on real data even when trained exclusively on synthetic data, suggesting they can bridge the synthetic to real domain gap.
For matting, SDMatte~\cite{sdmatte} applies diffusion models to interactive image matting, while GVM~\cite{GVM} focuses on portrait video matting with emphasis on hair details. However, these methods are limited to either single images or specific domains like human portraits, lacking the scale and generality needed for diverse video matting scenarios.


\paragraph{Pseudo-label generation in segmentation/matting } has become a key driver of recent progress. 
In the Segment Anything Model~\cite{sam, sam2}, the authors demonstrated that it is possible to create an iterative self-training loop by building a strong segmentation model and bootstrapping it with its own outputs.
ZIM~\cite{zim} trained a converter model that transforms image binary masks into alpha mattes using existing image matting datasets.

Inspired by the recent successes, we address video matting, which presents distinct challenges.
Unlike image matting~\cite{refmatte, am2k, p3m, DAViD} and video object segmentation, where abundant manually annotated ground-truth data~\cite{mose, sam2, davis, youtubevos} enables training strong labeler models, video matting datasets remain scarce and predominantly synthetic. 
This limitation hinders model generalization to real-world scenarios.
To overcome the challenge, we exploit the generative priors of pre-trained video diffusion models and train a robust pseudo-labeler using small-scale synthetic video matting annotations.

\section{Video Mask-to-Matte Model}
\label{sec:method}

\subsection{Problem Formulation}
Video matting is mathematically defined by the alpha compositing equation,
   $ I = \alpha F + (1 - \alpha) B, $
where $I$ is the observed image, $F$ is the foreground, $B$ is the background, and $\alpha \in [0,1]$ is the alpha matte representing pixel-level opacity. Unlike binary segmentation masks $M \in \{0, 1\}$, alpha mattes capture fine-grained details such as hair strands, motion blur, and intricate regions, which are critical for realistic video compositing.

There are several ways to specify the target object in video matting. 
One approach is to predefine the target semantic class, for example human portraits~\cite{p3m, GVM}, while another is to condition the model on additional signals such as points, boxes, or masks~\cite{sdmatte,maggie,mgmatting}.

We adopt binary segmentation masks as conditioning signals for two critical reasons: (1) masks allow the diffusion model to focus solely on generating fine-grained matting details (\emph{e.g.}, hair strands, motion blur) rather than inferring object boundaries, since shape information is already provided by the mask. 
This design decouples the challenges of object localization and matte generation, making the pipeline more robust and generalizable and
(2) Finally, since binary masks can be obtained through various sources, the applicability of a model can be greatly expanded. In this paper, we demonstrate that VideoMaMa can not only perform video matting in various settings when combined with existing video segmentation models, but also enables the conversion of existing video segmentation datasets into matting datasets. 
\begin{figure}[t]
    \centering
    \includegraphics[width=\columnwidth]{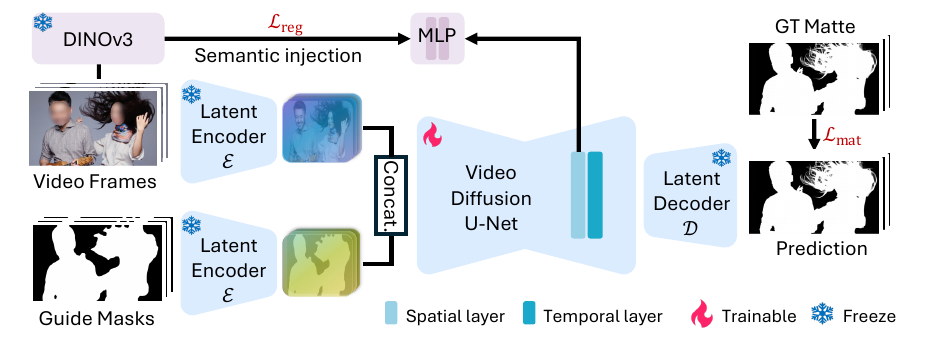}
    \vspace{-15pt}
    \caption{\textbf{Overview of VideoMaMa architecture.} RGB frames and guide masks are processed through video diffusion U-Net layers to generate high-quality video mattes. Semantic injection with DINO features is applied during training.}
    \vspace{-10pt}
    \label{fig:videomama}
\end{figure}

\subsection{Architecture Design}
\label{sec:videomama}



We build VideoMaMa on top of Stable Video Diffusion (SVD)~\cite{svd}, a pretrained video diffusion model originally designed for image-to-video generation. By leveraging generative priors, it can generate high-quality video matting annotations that exhibit realistic characteristics such as natural motion blur, proper edge detail, and temporal consistency, effectively transferring the model's learned understanding of video dynamics to the matting task.

We adapt the SVD architecture for mask-guided video matting by introducing mask-based conditioning in the latent space and employing single-step inference for improved efficiency. In addition, we propose a semantic knowledge injection technique that enhances the model’s understanding of object boundaries and strengthens temporal consistency when tracking complex objects with fine structural details.
An overview of the model is shown in Figure~\ref{fig:videomama}.

\paragraph{Latent space formulation.}
Following modern video diffusion models, we operate in a compressed latent space to alleviate the computational overhead of processing high resolution videos. We leverage a Variational Autoencoder (VAE)~\cite{vae} for efficient encoding and decoding. All inputs, including video frames $V = \{I_t\}_{t=1}^{T}$, binary mask track $M = \{M_t\}_{t=1}^{T}$, and alpha mattes $\alpha = \{\alpha_t\}_{t=1}^{T}$, are encoded into the same latent space:
    $z_x = \mathcal{E}(x), \hat{x} = \mathcal{D}(z_x),$
where $\mathcal{E}$ and $\mathcal{D}$ denote the encoder and decoder of the VAE, respectively, and $x$ may represent $V$, $M$, or $\alpha$. Since all three modalities share the same spatial dimensions, this unified encoding approach naturally allows them to be processed together in the compressed latent space, reducing memory requirements while preserving spatial correspondence.

\paragraph{Single-step diffusion formulation.}
We formulate VideoMaMa as a single-timestep generative model that directly synthesizes high-fidelity alpha matte latents $\hat{z}_{\alpha}$ in a single forward pass, where $h$ and $w$ represent the compressed spatial dimensions and $c$ is the latent channel dimension. Unlike traditional diffusion models that require iterative denoising steps, our model performs direct prediction from noise to clean latents, making it significantly more efficient for large-scale dataset generation. 

The generation process takes as input the frame-wise concatenation of video latents $z_V$, mask latents $z_M$, and Gaussian noise $\epsilon \sim \mathcal{N}(0, I)$ along the channel dimension:
\begin{equation}
    \hat{z}_{\alpha} = \mathcal{F}_\text{SVD}(\text{concat}(z_V, z_M, \epsilon)),
    \label{eq:single_step_model}
\end{equation}
where $\mathcal{F}_\text{SVD}$ represents our adapted SVD model, and $\text{concat}(\cdot)$ denotes a concatenation. We modify SVD's architecture by replacing the original image conditioning input with this concatenated tensor, allowing the model to leverage SVD's strong temporal modeling capabilities while conditioning on both appearance (from $z_V$) and shape information (from $z_M$). The final alpha matte video is obtained by decoding the predicted latents through the VAE decoder: $\hat{\alpha} = \mathcal{D}(\hat{z}_{\alpha})$.

\subsection{Training Recipe}
\paragraph{Mask augmentation.}
While providing binary masks as input conditioning signals simplifies the task for the diffusion model, it also introduces the risk of copy-paste behavior. For objects with simple shapes (e.g., cars or desks) or masks that already contain fine-grained details (e.g., binarized alpha mattes), the model may trivially copy the mask instead of generating realistic matting details by reasoning about the RGB image.

\begin{figure}[t!]
    \centering
    \includegraphics[width=\linewidth]{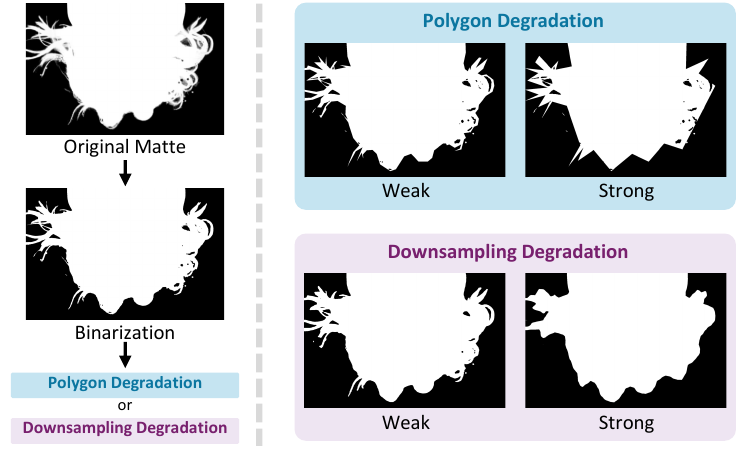}
    \vspace{-20pt}
    \caption{\textbf{Examples of mask augmentation methods.} Polygon and Downsampling degradation are applied at weak and strong augmentation levels.}
    \vspace{-10pt}
    \label{fig:augmentation}
\end{figure}

To prevent this, we apply mask augmentation during training to remove fine-grained information from the input masks, forcing the model to infer matting details from the RGB image.
Our mask augmentation strategy consists of two operations (see Figure~\ref{fig:augmentation}): (1) Polygon Degradation: We approximate the mask boundary with a polygon, simplifying the contour and removing fine details. (2) Downsampling Degradation: We downsample and upsample the mask at the same scale, which effectively removes high-frequency details while preserving the overall shape. These augmentations create a gap between the coarse input mask and the fine-grained target alpha matte, encouraging the model to leverage appearance cues from the RGB video to generate realistic matting details.

\paragraph{Two-stage training.}
For matting tasks, both training and inference resolution are critical. As matting operates at the pixel level with continuous values, downsampling to lower resolutions can degrade fine-grained matting details. However, training video diffusion models at high resolution is computationally prohibitive. To address this challenge, we propose a two-stage training strategy that decomposes the learning process: first training spatial layers at high resolution on single frames to capture fine details, then training temporal layers at lower resolution on video sequences to learn temporal consistency.
In the first stage, we freeze the temporal layers of SVD and train only the spatial layers, enabling the model to focus on capturing pixel-level details at high resolution. While this stage allows the model to learn fine-grained features within individual frames, it lacks temporal consistency across video sequences. In the second stage, we freeze the spatial layers to preserve the learned detail-capturing capability and fine-tune only the temporal layers to learn temporal coherence and video-specific characteristics. This decomposed training strategy enables our model to achieve both high-resolution detail and temporal consistency without requiring full high-resolution video training.


\paragraph{Matting loss function.}
We adopt v-parameterization~\cite{v_param} for single-step generation, where the model is trained to directly convert noise into clean alpha matte latents. The predicted latent $\hat{z}_{\alpha}$ is decoded to pixel space $\hat{\alpha} = \mathcal{D}(\hat{z}_{\alpha})$, and the loss is computed at the pixel level:
\begin{equation}
    \mathcal{L}_{\text{mat}} = \mathbb{E}_{\alpha, z_V, z_M, \epsilon} \left[ \text{sim}(\mathcal{D}(\hat{z}_{\alpha}), \alpha) \right],
    \label{eq:single_step_loss}
\end{equation}
where $\text{sim}(\cdot, \cdot)$ comprises pixel-wise similarity loss terms between the decoded prediction and ground-truth alpha matte. This pixel-level supervision enables direct optimization for visual quality, capturing fine-grained details such as motion blur and intricate boundary structures.

\paragraph{Semantic knowledge injection.}
While diffusion priors are powerful for generating precise alpha mattes, they struggle with semantic understanding of object boundaries and consistently tracking objects, particularly for complex objects with intricate structures. To address this, we inject semantic knowledge into VideoMaMa by aligning SVD features with robust semantic representations from DINOv3~\cite{dinov3} throughout training. 

We first extract DINO features $h_{\text{dino}} = \mathcal{F}_{\text{dino}}(V)$ from the video frames $V$. Then, we extract intermediate features $h^l$ from the $l$-th layer of the diffusion model and project them into the DINO feature space using a learnable MLP $p_{\phi}$ with parameters $\phi$. We minimize the alignment loss that maximizes patch-wise cosine similarities:
\begin{equation}
\mathcal{L}_{\text{reg}} = -\mathbb{E}_{\alpha, z_V, z_M, \epsilon,V} \Big[ \text{cos-sim}(h_{\text{dino}}, p_{\phi}(h^{l})) \Big],
\end{equation}
where $n$ indexes patches and $\text{cos-sim}(\cdot, \cdot)$ denotes cosine similarity. This design allows the model to leverage semantic understanding of object categories and structures while maintaining the generative capabilities of the diffusion backbone, enabling more accurate matte generation for challenging cases such as overlapping objects or complex articulated structures.
\section{Matte Anything in Videos (\mbox{MA-V}) Dataset}
\label{sec:mav}
\begin{figure*}[!t]
    \centering
    \includegraphics[width=0.95\linewidth]{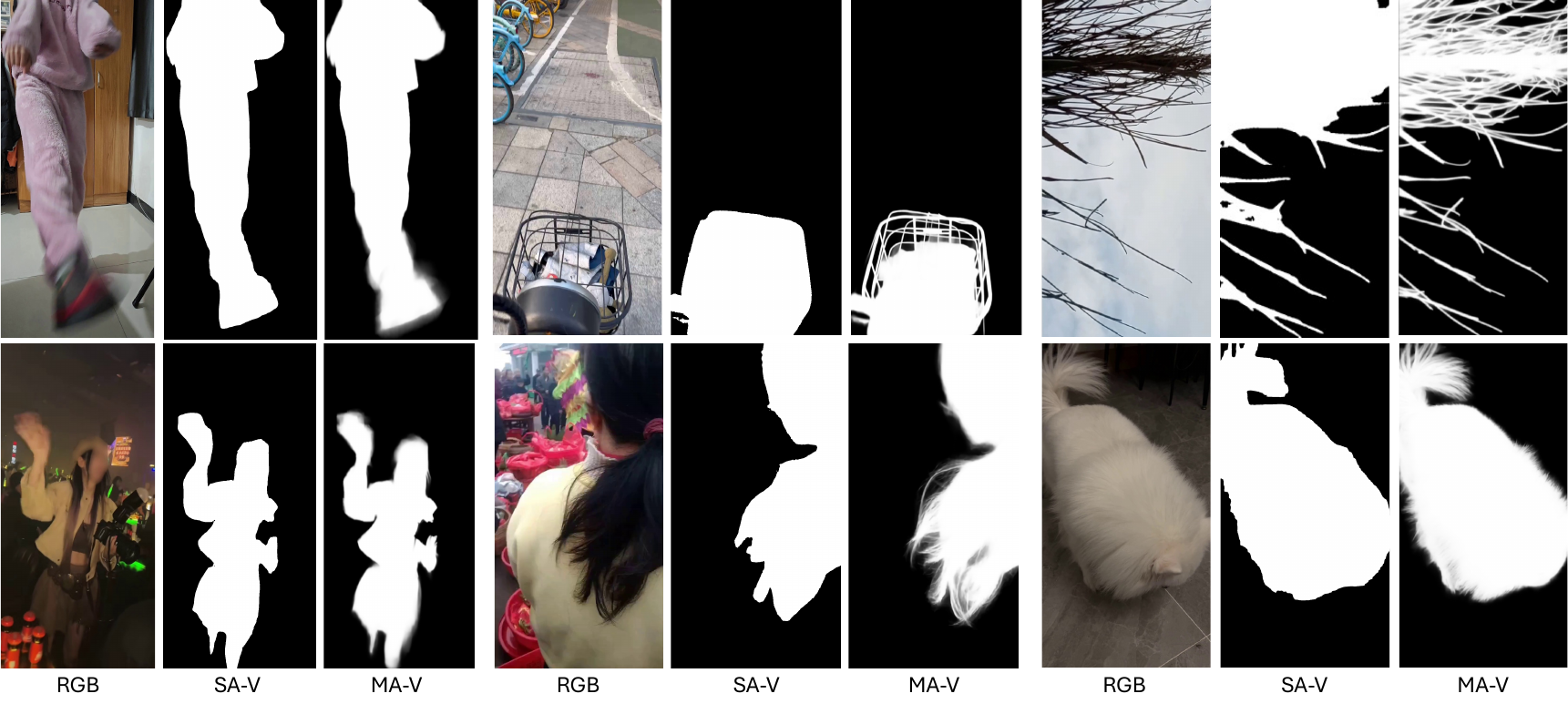} 
    \vspace{-15pt}
    \caption{\textbf{Qualitative examples from our MA-V dataset.} We show RGB frames with our high-quality MA-V annotations and original SA-V~\cite{sam2} masks for comparison. MA-V provides refined alpha mattes for diverse scenarios.
    }
    \label{fig:mav_sav_comparison}
    \vspace{-5pt}
\end{figure*}

\begin{table}[t]
    \centering
    \small
    \caption{\textbf{Dataset statistics for video matting.}}
    \vspace{-5pt}
    \label{tab:dataset_stats}
    \resizebox{\columnwidth}{!}{%
    \begin{tabular}{l c r c} 
        \toprule
        \textbf{Dataset} & \textbf{Category} & \textbf{\# Videos} & \textbf{Composition Required} \\ 
        \midrule
        SynHairMan~\cite{GVM} & Human & 200 & Yes \\ 
        VideoMatte240K~\cite{video240k} & Human & 478 & Yes \\ 
        V-HIM2K5~\cite{maggie} & Human & 443 & Yes \\ 
        VideoMatting108~\cite{video108} & Various & 108 & Yes \\ 
        CRGNN~\cite{crgnn} & Various & 60 & Yes \\ 
        VM800~\cite{matanyone} & Human & 826 & Yes \\ 
        \midrule
        \textbf{MA-V (Ours)} & \textbf{Various} & \textbf{50,541} & \textbf{No} \\ 
        \bottomrule
    \end{tabular}
    }
    \vspace{-10pt}
\end{table}


The difficulty of obtaining video matting annotations has severely limited existing datasets. As shown in Table~\ref{tab:dataset_stats}, prior datasets contain at most hundreds of videos, predominantly focusing on human subjects captured in controlled environments~\cite{video240k,video108} or through manual annotation~\cite{p3m, sparsemat, Distinctions646, Aim}. While composition-based approaches enable scaling and precise annotations, they create artificial scene compositions that differ fundamentally from natural video footage, limiting model generalization to real world scenarios.

\subsection{Dataset Construction} To address these limitations, we introduce Matte Anything in Videos (\mbox{MA-V}), created by applying VideoMaMa to annotate SA-V's~\cite{sam2} diverse mask annotations. As shown in Table~\ref{tab:dataset_stats}, \mbox{MA-V} provides 50,541 videos captured in natural settings, nearly 50× larger than existing real-video datasets. Unlike prior work that focuses predominantly on humans, \mbox{MA-V} encompasses diverse object categories at multiple scales, from small objects to full scenes. 
The dataset construction process is straightforward yet effective: we leverage VideoMaMa's ability to generate high-quality alpha mattes from binary masks, converting SA-V's segmentation annotations into continuous matting labels while preserving the natural scene context. Critically, \mbox{MA-V} is the first large-scale video matting dataset where both foreground and background naturally co-occur in real captured footage, eliminating the synthetic composition gap that limits previous datasets. Figure~\ref{fig:mav_sav_comparison} illustrates the visual quality and diversity of \mbox{MA-V} compared to SA-V's binary masks, demonstrating how VideoMaMa generates fine-grained matting details including semi-transparent regions, motion blur, and intricate boundary structures regardless of the target category that are absent in the source segmentation masks.

\subsection{Fine-tuning SAM2 with MA-V}
We finetune SAM2~\cite{sam2} on our MA-V dataset, referred to as \textbf{SAM2-Matte} hereafter. SAM2, originally designed for binary segmentation, requires minimal adaptation for matting: we apply a sigmoid function after the mask logits to produce continuous alpha values in the range [0,1]. To assess \mbox{MA-V} quality, we also finetune SAM2 on existing video matting datasets and evaluate their performance on standard matting benchmarks.
\section{Experiments} 
\label{sec:exp}
\begin{figure*}[t]
    \centering
    \includegraphics[width=\textwidth]{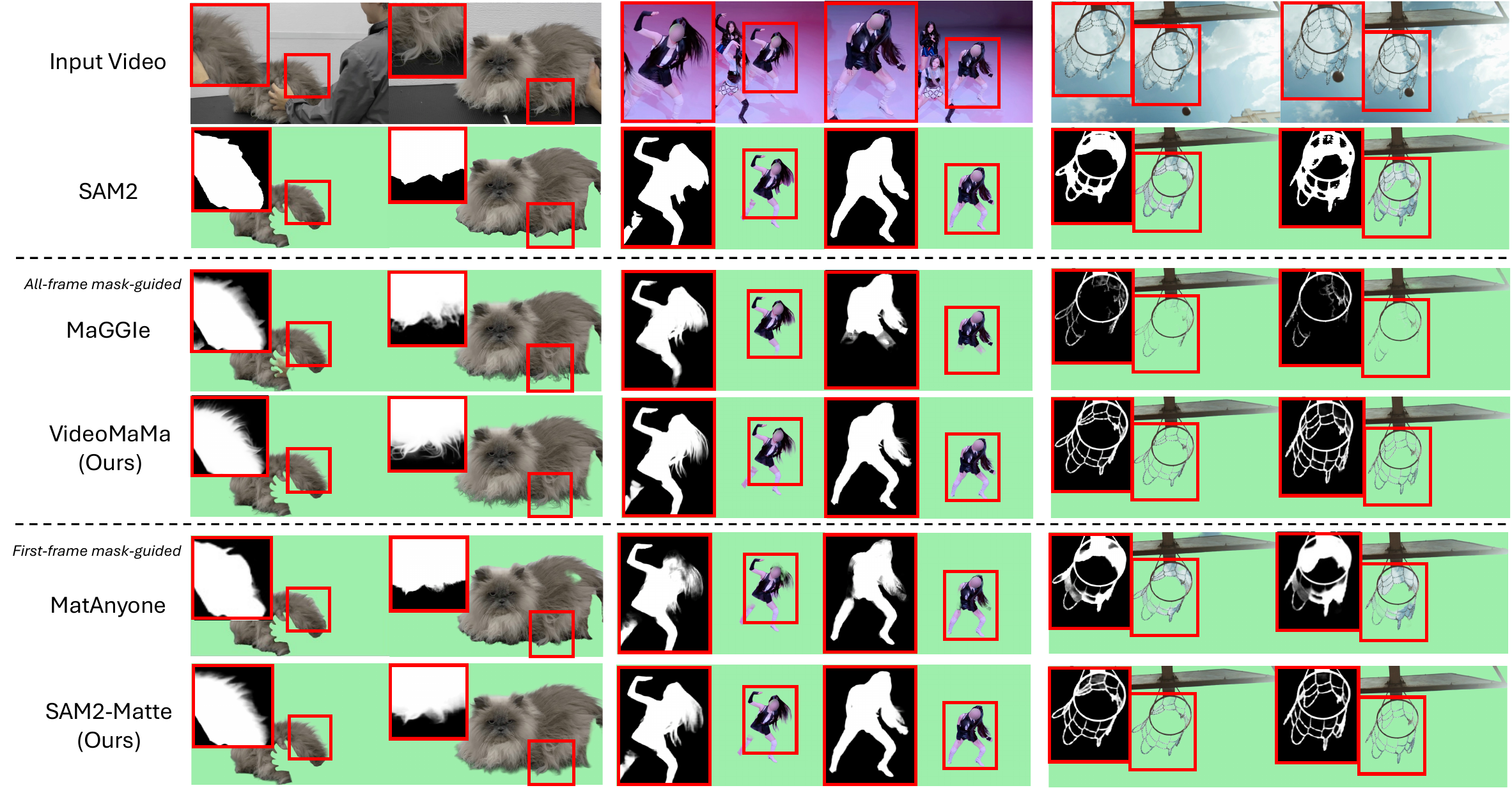}
    \vspace{-15pt}
    \caption{
    \textbf{Qualitative comparison on in-the-wild videos.} 
        We evaluate two settings: (1) all-frame mask-guided video matting where VideoMaMa is compared against MaGGIe~\cite{maggie}, and (2) first-frame mask-guided matting where SAM2-Matte is compared against MatAnyone~\cite{matanyone}. All methods use SAM2~\cite{sam2} to generate mask inputs. 
    }
    \vspace{-15pt}
    \label{fig:videomama_qual}
\end{figure*}
\begin{table*}[t]
  \centering
   \caption{\textbf{All-frame mask-guided video matting comparison on V-HIM60 and YouTubeMatte benchmarks.} 
    We compare VideoMaMa (Ours) against mask-guided matting methods: MaGGIe~\cite{maggie} (video mask-guided) and MGM~\cite{mgmatting} (image mask-guided). We evaluate on two mask types: \textit{Synthetically Degraded Masks} including downsampling (8$\times$, 32$\times$) and polygon degradation with varying difficulty levels, and \textit{Model-Generated Masks} from SAM2~\cite{sam2}. Lower values indicate better performance.
    }
    \vspace{-2mm}
  \label{tab:refinement_benchmark_wide}
  \resizebox{\textwidth}{!}{%
  \begin{tabular}{@{}lcccccccccccccccc@{}}
    \toprule
    & \multicolumn{8}{c}{\textbf{V-HIM60~\cite{maggie} (Hard)}} & \multicolumn{8}{c}{\textbf{YouTubeMatte~\cite{matanyone} (1920 $\times$ 1080)}} \\
    \cmidrule(lr){2-9} \cmidrule(lr){10-17}
    & \multicolumn{4}{c}{\textbf{MAD $\downarrow$}} & \multicolumn{4}{c}{\textbf{Gradient $\downarrow$}} & \multicolumn{4}{c}{\textbf{MAD $\downarrow$}} & \multicolumn{4}{c}{\textbf{Gradient $\downarrow$}} \\
    \cmidrule(lr){2-5} \cmidrule(lr){6-9} \cmidrule(lr){10-13} \cmidrule(lr){14-17}
    \textbf{Input Mask} & \textbf{Input} & \textbf{MGM~\cite{mgmatting}} & \textbf{MaGGle~\cite{maggie}} & \textbf{Ours} & \textbf{Input} & \textbf{MGM~\cite{mgmatting}} & \textbf{MaGGle~\cite{maggie}} & \textbf{Ours} & \textbf{Input} & \textbf{MGM~\cite{mgmatting}} & \textbf{MaGGle~\cite{maggie}} & \textbf{Ours} & \textbf{Input} & \textbf{MGM~\cite{mgmatting}} & \textbf{MaGGle~\cite{maggie}} & \textbf{Ours} \\
    \midrule
    \multicolumn{17}{@{}l}{\textit{Synthetically Degraded Masks}} \\
    \midrule
    Downsampl. 8x          & 2.744 & 63.8374 & 2.3790 & \textbf{1.306} & 11.127 & 16.5609 & 5.4150 & \textbf{2.363} & 1.946 & 3.4973 & 1.6420 & \textbf{0.934} & 8.034 & 4.0822 & 4.9733 & \textbf{2.031} \\
    Downsampl. 32x        & 5.132 & 79.2273 & 2.5375 & \textbf{1.461} & 33.664 & 22.2024 & 5.9073 & \textbf{2.849} & 4.203 & 3.9061 & 1.8171 & \textbf{1.029} & 36.540 & 5.3603 & 5.6318 & \textbf{2.560} \\
    Polygon. (Easy)           & 3.636 & 64.5457 & 2.5808 & \textbf{1.416} & 16.000 & 17.3675 & 5.5572 & \textbf{2.480} & 3.619 & 4.1390 & 2.2250 & \textbf{1.206} & 19.155 & 4.8579 & 5.6773 & \textbf{2.335} \\
    Polygon. (Hard)           & 6.771 & 75.5501 & 3.1567 & \textbf{1.640} & 43.228 & 20.5516 & 6.4121 & \textbf{3.024} & 9.470 & 4.4549 & 2.3111 & \textbf{1.404} & 65.844 & 5.6132 & 5.9184 & \textbf{3.237} \\
    \midrule
    \multicolumn{17}{@{}l}{\textit{Model-Generated Masks}} \\
    \midrule
    SAM2~\cite{sam2}           & 4.666 & 66.5221 & 3.1567 & \textbf{2.435} & 21.960 & 18.2103 & 6.4121 & \textbf{4.180} & 3.569 & 4.1136 & 1.9499 & \textbf{1.737} & 24.729 & 5.0448 & 5.7877 & \textbf{3.635} \\
    \bottomrule
  \end{tabular}
  }
  \vspace{-10pt}
\end{table*}

\subsection{Experimental Details}
\paragraph{Implementation details.}
We train VideoMaMa based on the Stable Video Diffusion~\cite{svd} (SVD) model using diverse image and video matting datasets. Our two-stage training strategy employs different resolutions optimized for each stage's objective. In stage 1, we train the spatial layers of SVD on single images at high resolution (1024×1024) to capture fine-grained matting details such as hair strands and intricate boundary structures. We freeze the temporal layers during this stage to focus learning on spatial precision. In stage 2, we freeze the learned spatial layers and train only the temporal layers on video sequences. To balance computational efficiency with temporal modeling, we use 3-frame clips at 704×704 resolution, which enables the model to learn temporal consistency and motion-aware matting characteristics. To inject semantic knowledge, we incorporate features from a frozen DINOv3~\cite{dinov3} encoder, aligning the DINO features with the first upsampling block of the SVD decoder. This integration provides the diffusion model with semantic understanding of object structures while preserving its generative capabilities.

\paragraph{Training details.}
For both stages, we use a batch size of 64 and a learning rate of $5 \times 10^{-5}$ with the AdamW optimizer. The training objective, $\mathcal{L}_{\text{mat}}$, combines L1 loss for pixel-wise accuracy and Laplacian loss for preserving edge sharpness and boundary details. Each stage is trained for 10,000 iterations until convergence. All experiments are conducted on NVIDIA A100 GPUs with mixed-precision training for efficiency.
SAM2-Matte is trained on the combination of existing video matting datasets and \mbox{MA-V}. Details of the existing datasets are provided in the appendix.

\subsection{Quantitative Evaluations}

\paragraph{All-frame mask-guided video matting.}
We evaluate VideoMaMa following the protocol in~\cite{maggie}, where input binary masks for every frame are provided. We use two categories of input masks: (1) Synthetically Degraded Masks through downsampling (8$\times$, 32$\times$) and polygon degradation (easy, hard), mimicking imperfect user annotations, and (2) Model-Generated Masks from SAM2~\cite{sam2}, representing automatic segmentation pipelines. This diverse set evaluates VideoMaMa's robustness across different mask quality levels.

In Table~\ref{tab:refinement_benchmark_wide}, we compare VideoMaMa against MaGGIe~\cite{maggie} and MGM~\cite{mgmatting}. For MGM, which is an image-based method, we perform inference on each frame independently. We report metrics comparing both input masks and generated results against ground-truth alpha mattes. All evaluations use 12-frame sequences.

We adopt standard metrics: MAD for overall accuracy and Gradient error~\cite{rhemann2009perceptually} for boundary quality. Results are reported on V-HIM60 Hard~\cite{maggie} and YouTubeMatte 1920$\times$1080~\cite{matanyone}.

VideoMaMa consistently outperforms existing mask-guided matting models across both synthetically degraded and model-generated masks. Our evaluation across different mask styles demonstrates that VideoMaMa can handle various types of mask inputs and produce high-quality video mattes consistently.

\begin{table*}[t]
  \centering
  \caption{\textbf{First-frame mask-guided video matting comparison on V-HIM60 and YouTubeMatte benchmarks.} 
    We evaluate video matting performance across different difficulty levels in V-HIM60. The \textbf{best} and \underline{second-best} results are highlighted.
    }
    \vspace{-3mm}
  \label{tab:first_mgm_quan}
  \resizebox{\textwidth}{!}{%
  \begin{tabular}{@{}l cccc cccc cccc cccc@{}}
    \toprule
    \multirow{3}{*}{\textbf{Method}} & \multicolumn{12}{c}{\textbf{V-HIM60~\cite{maggie}}} & \multicolumn{4}{c}{\textbf{YouTubeMatte~\cite{matanyone}}} \\
    \cmidrule(lr){2-13} \cmidrule(lr){14-17}
    & \multicolumn{4}{c}{\textbf{Easy}} & \multicolumn{4}{c}{\textbf{Medium}} & \multicolumn{4}{c}{\textbf{Hard}} & \multicolumn{4}{c}{\textbf{1920 $\times$ 1080}} \\
    \cmidrule(lr){2-5} \cmidrule(lr){6-9} \cmidrule(lr){10-13} \cmidrule(lr){14-17}
    & MAD & MAD-T & MSE & GRAD & MAD & MAD-T & MSE & GRAD & MAD & MAD-T & MSE & GRAD & MAD & MAD-T & MSE & GRAD \\
    \midrule
    SAM2~\cite{sam2} & 2.7978 & 91.5246 & 1.4799 & 13.3307 & 4.9618 & 105.2547 & 2.9401 & 25.6735 & 7.8517 & 130.0338 & 6.0075 & 26.3355 & 3.8530 & 103.2181 & 2.6133 & 37.4074 \\
    MatAnyone~\cite{matanyone} & 3.5803 & 92.8273 & 1.9905 & 7.0763 & 7.1955 & 100.0465 & 4.7586 & 11.4541 & 5.7195 & 102.4861 & 3.4620 & 9.8199 & 1.9909 & 51.0939 & 0.7085 & 8.9019 \\
    SAM2+VideoMaMa & \underline{1.3539} & \textbf{40.6830} & \underline{0.4063} & \textbf{2.8876} & \underline{2.3132} & \textbf{49.0329} & \underline{0.9335} & \textbf{4.6944} & \underline{2.7757} & \textbf{53.4392} & \underline{1.4428} & \textbf{4.8276} & \underline{1.3559} & \underline{35.5064} & \underline{0.2940} & \underline{3.5742} \\
    SAM2-Matte & \textbf{1.3446} & \underline{49.0329} & \textbf{0.2932} & \underline{3.1245} & \textbf{2.2710} & \underline{51.8996} & \textbf{0.6669} & \underline{5.2227} & \textbf{2.6112} & \underline{58.7796} & \textbf{1.0750} & \underline{5.0869} & \textbf{1.2695} & \textbf{33.5268} & \textbf{0.2831} & \textbf{3.4640} \\
    \bottomrule
  \end{tabular}
  } 
  \vspace{-10pt}
\end{table*}
\paragraph{First-frame mask-guided video matting.}
In Table~\ref{tab:first_mgm_quan}, we evaluate VideoMaMa and SAM2-Matte following a setting proposed in~\cite{matanyone} where a binary mask guidance is provided only for the first frame. For VideoMaMa, the binary mask guidance across all frames is generated by propagating the mask from the first frame using SAM2~\cite{sam2}, denoted as SAM2+VideoMaMa. We also report the score of the binary masks produced by SAM2 to provide a baseline for the improvements achieved by VideoMaMa.

We employ multiple evaluation metrics: MAD for overall semantic accuracy, MSE for pixel-level error, MAD-T (trimap-based MAD) which computes MAD only within the uncertain trimap region derived from the ground-truth alpha map, and Gradient error~\cite{rhemann2009perceptually} for assessing boundary detail extraction quality. The trimap-based evaluation is particularly important as it focuses on the challenging boundary regions where precise matting is most critical. 

SAM2-Matte trained on the combined dataset surpasses previous state-of-the-art methods, MatAnyone~\cite{matanyone}, on challenging benchmarks: V-HIM60 Hard~\cite{maggie} and YouTubeMatte~\cite{matanyone}. This demonstrates that \mbox{MA-V} provides high-quality training data for video matting models, enabling to extract matting quality and temporal consistency over the video frames. Note that, for VideoMaMa in this experiment, we processed each of the 12-frame segments and merged the results to obtain the complete video matting.

\subsection{Qualitative Evaluations}
To further evaluate the model’s performance on real-world videos, we present qualitative comparisons in Figure~\ref{fig:videomama_qual}. We categorize the methods into two settings: first-frame mask-guided and all-frame mask-guided. The initial binary masks are obtained manually using the SAM2 model. Unlike previous matting methods~\cite{matanyone, maggie}, VideoMaMa and SAM2-Matte not only perform well on human portraits but also generalize to a wide range of object categories and in-the-wild video content. Moreover, the high-quality results from SAM2-Matte demonstrate that \mbox{MA-V}’s large-scale and diverse annotations effectively enhance video matting performance beyond what existing datasets can achieve.

\subsection{Ablation Studies}

\begin{table}[t]
  \centering
    \caption{\textbf{Ablation study on the number of frame during inference.} We report the mask refinement performance in MAD metric on YouTubeMatte dataset~\cite{matanyone} with different input masks. }
    \vspace{-5pt}
  \label{tab:num_frames}
  \resizebox{\columnwidth}{!}{%
  \begin{tabular}{@{}llccccc@{}}
    \toprule
    \multicolumn{2}{c}{} & \multicolumn{5}{c}{\textbf{Number of Frames per Inference}} \\
    \cmidrule(lr){3-7}
    \textbf{Mask Input} & & \textbf{24} & \textbf{18} & \textbf{12} & \textbf{6} & \textbf{1} \\
    \midrule
    \multirow{2}{*}{Downsampl. 32x} & Input & 4.2411 & 4.2455 & 4.2029 & 4.1377 & 4.1336 \\
    & Refined & 1.0237 & 1.0272 & 1.0292 & 1.0418 & 1.1024 \\
    \midrule
    \multirow{2}{*}{Polygon. (Hard)} & Input & 9.5559 & 9.6224 & 9.4700 & 8.7498 & 8.5455 \\
    & Refined & 1.4472 & 1.4847 & 1.4042 & 1.3382 & 1.5598 \\
    \midrule
    \multirow{2}{*}{SAM2~\cite{sam2}} & Input & 3.6023 & 3.5991 & 3.5693 & 3.5476 & 3.8711 \\
    & Refined & 1.6921 & 1.7273 & 1.7370 & 1.7948 & 1.8828 \\
    \bottomrule
  \end{tabular}%
  }
\end{table}
\paragraph{Number of inference frames.}
We evaluate VideoMaMa's performance across varying numbers of frames. Although trained on maximum 3 frames, VideoMaMa successfully handles diverse frame counts from single frames to 24 frames at inference time. As shown in Table~\ref{tab:num_frames}, the model maintains consistent performance across different frame counts, demonstrating strong temporal generalization. 

\begin{table}[t]
  \centering
  \caption{\textbf{Ablation study on training recipe.} We evaluate MAD with different training configurations on YouTubeMatte~\cite{matanyone}. Scores for input masks are shown below the each input mask type. Percentages in parentheses indicate the relative improvement (\textcolor{green!70!black}{green}) or degradation (\textcolor{red!80!black}{red}) compared to the input score.}
  \label{tab:stage_wise}
  \resizebox{\columnwidth}{!}{%
  \begin{tabular}{@{}ccc|ccc@{}}
    \toprule
    \textbf{S1} & \textbf{S2} & \textbf{DINO} & \makecell{\textbf{Downsampl. 32x} \\ {Input: 4.2029}} & \makecell{\textbf{Polygon. (Hard)} \\ {Input: 9.4700}} & \makecell{\textbf{SAM2} \\ {Input: 3.5693}} \\
    \midrule
    \cmark & \xmark & \xmark & 3.7620 (\textcolor{green!70!black}{10.49\%}) & 4.8306 (\textcolor{green!70!black}{49.00\%}) & 4.0496 (\textcolor{red!80!black}{-13.46\%}) \\
    \xmark & \cmark & \xmark & 1.2350 (\textcolor{green!70!black}{70.62\%}) & 2.2372 (\textcolor{green!70!black}{76.38\%}) & 2.3124 (\textcolor{green!70!black}{35.21\%}) \\
    \cmark & \cmark & \xmark & 1.2552 (\textcolor{green!70!black}{70.13\%}) & 1.9975 (\textcolor{green!70!black}{78.91\%}) & 1.9404 (\textcolor{green!70!black}{45.64\%}) \\
    \cmark & \cmark & \cmark & 1.0292 (\textcolor{green!70!black}{75.51\%}) & 1.4042 (\textcolor{green!70!black}{85.17\%}) & 1.7370 (\textcolor{green!70!black}{51.34\%}) \\
    \bottomrule
  \end{tabular}
  } 
\end{table}
\paragraph{Training recipe.}
We ablate our two-stage training strategy and semantic knowledge injection on YouTubeMatte~\cite{matanyone} with four configurations: (1) image only (stage 1), (2) video only (stage 2), (3) two-stage without semantic knowledge, and (4) two-stage training with DINO features. As shown in Table~\ref{tab:stage_wise}, having both stage 1 and stage 2 training stages led to the best results. Adding DINO semantic features further improves performance enabling better object understanding and boundary localization. The results validate that both components are necessary and complementary for optimal performance.

\begin{table}[t]
\centering
\caption{
    \textbf{Ablation study for training data} on V-HIM60 Hard~\cite{maggie} and DAVIS val~\cite{davis},
    including (a) ED, (B) MAV, (c) ED + MA-V.
    ED denotes that existing dataset.
}
\label{tab:ablation_combined_matting_davis}
\resizebox{\columnwidth}{!}{%
\renewcommand{\arraystretch}{0.9} 
\begin{tabular}{l
                S[table-format=1.4]
                S[table-format=1.4]
                S[table-format=1.4]
                S[table-format=2.1]
                S[table-format=2.1]
                S[table-format=2.1]}
\toprule
\multirow{2}{*}{\textbf{Method}} & \multicolumn{3}{c}{\textbf{V-HIM60 (Hard)}} & \multicolumn{3}{c}{\textbf{DAVIS}} \\
\cmidrule(lr){2-4} \cmidrule(lr){5-7} 
& {MAD} & {MSE} & {GRAD} & {J\&F} & {J} & {F} \\
\midrule
MatAnyone~\cite{matanyone} & 4.6655 & 2.5870 & 8.8870 & 79.7 & 76.5 & 82.9 \\
\midrule
(a) ED & 7.5798 & 5.6824 & 8.7091 & 77.0 & 74.1 & 80.0 \\
(b) MA-V & 3.1830 & 1.5603 & 5.7869 & \textbf{87.9} & \textbf{85.2} & \textbf{90.6} \\
(c) ED + MA-V & \textbf{2.6112} & \textbf{1.0750} & \textbf{5.0869} & 85.9 & 83.5 & 88.3 \\
\bottomrule
\end{tabular}%
\renewcommand{\arraystretch}{1.0} 
} 
\vspace{-10pt}
\end{table}
\begin{figure}
    \centering
    \includegraphics[width=\columnwidth]{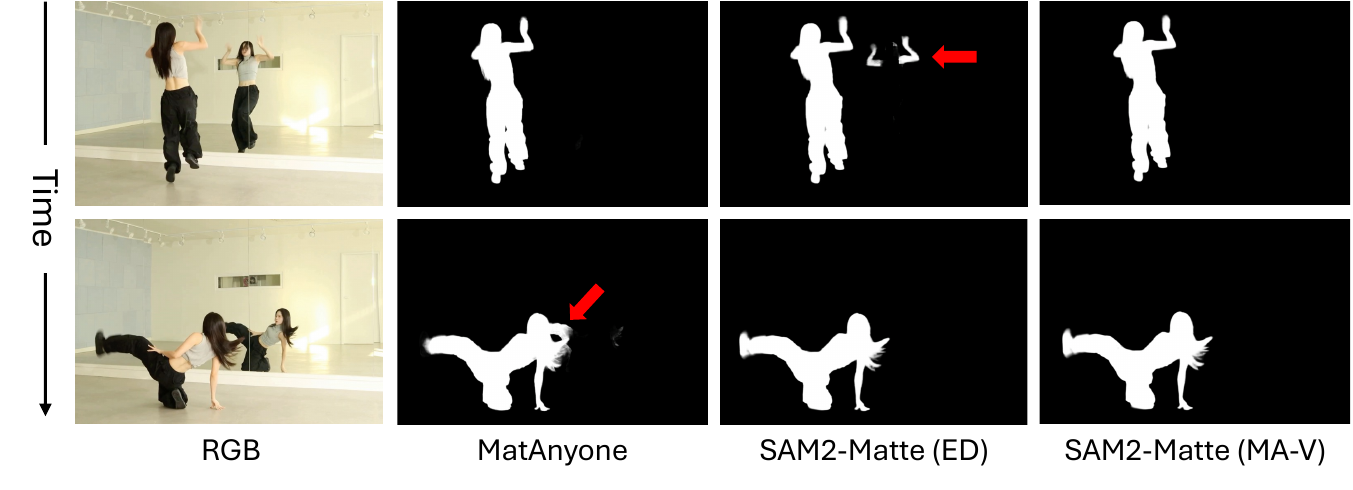}
    \vspace{-20pt}
    \caption{
    \textbf{Comparison on a real video between MatAnyone~\cite{matanyone} and SAM2-Matte variants.}
    Red arrows highlight regions where our method achieves the best results.
    } 
    \label{fig:sam2-matte_qual}
    \vspace{-15pt}
\end{figure}

\paragraph{Impact of MA-V Dataset.}
We assess the impact of MA-V on matting and tracking performance on first-frame mask-guided matting task. We evaluate SAM2-Matte trained under different data configurations by comparing (a) existing datasets only (ED), (b) \mbox{MA-V} only, and (c) both combined. A detailed list of the existing matting datasets used in this experiment will be included in the appendix.

\label{pargraph:mav_analysis}
To validate the impact of \mbox{MA-V} on matting task, we report matting quality metrics on the V-HIM60 (Hard)~\cite{maggie}. As shown in Table~\ref{tab:ablation_combined_matting_davis}, Configuration (a) shows limited performance due to the small scale and lack of diversity in existing datasets. Remarkably, (b) significantly outperforms (a) and even surpasses state-of-the-art results, demonstrating the effectiveness of \mbox{MA-V}. When combining the existing matting datasets with \mbox{MA-V} in (c), we achieve the best overall matting quality. We hypothesize that the shared synthetic nature of both the existing matting datasets and V-HIM60 contributes to this improved performance.

We evaluate tracking performance on DAVIS~\cite{davis} using VOS metrics by binarizing matting results. As shown in Table~\ref{tab:ablation_combined_matting_davis}, configuration (b) substantially improves performance over (a), demonstrating the effectiveness of \mbox{MA-V}. Interestingly, adding the existing matting datasets in configuration (c) leads to degraded tracking performance compared to using \mbox{MA-V} alone, suggesting that existing synthetic datasets introduce domain biases that reduce tracking robustness on in-the-wild videos.

In Figure~\ref{fig:sam2-matte_qual}, we compare MatAnyone~\cite{matanyone} and SAM2-Matte variants. The results show that using MA-V enables robust tracking even along ambiguous boundaries.
\section{Conclusion}
\label{sec:conclusion}

We introduce VideoMaMa, a diffusion-based model that leverages generative priors to perform robust video matting. By exploiting VideoMaMa's strong generalization capability, we construct \mbox{MA-V}, the first large-scale pseudo video matting dataset built on real-world videos, by converting segmentation masks from SA-V. Through comprehensive experiments, we demonstrate that both VideoMaMa and models trained on \mbox{MA-V} (e.g., SAM2-Matte) achieve state-of-the-art performance across diverse video matting benchmarks, validating the effectiveness of our data generation approach. We believe that VideoMaMa and \mbox{MA-V} will significantly contribute to advancing video matting research by providing both a VideoMaMa and a large-scale, high-quality dataset that bridges the synthetic-to-real domain gap.


\setcounter{section}{0}
\renewcommand{\thesection}{\Alph{section}}

\section*{\Large Appendix}
This supplementary material provides additional details on implementation, evaluation metrics, and qualitative results that could not be included in the main paper due to space constraints. 


The document is organized as follows:

\begin{itemize}
    
    \item \textbf{Section~\ref{sec:additional_implementation}} elaborates on the training configurations for SAM2-Matte and details the architectural design of the VideoMaMa feature injection module.
    
    \item \textbf{Section~\ref{sec:MAD-T}} provides the formal definition and algorithm for the Trimap-based Mean Absolute Difference (MAD-T) metric, designed to evaluate transition region accuracy frame-by-frame.
    
    \item \textbf{Section~\ref{sec:sam2_comparison}} presents a comparative analysis between SAM2-Matte and the original binary SAM2, empirically demonstrating the necessity and effectiveness of the \mbox{MA-V} dataset.
    
    \item \textbf{Section~\ref{sec:limitation}} analyzes the limitations of VideoMaMa, specifically discussing failure cases arising from reliance on semantically inaccurate instance masks.
\end{itemize}




\section{Additional Implementation Details}
\label{sec:additional_implementation}
\paragraph{VideoMaMa}
To obtain in-the-wild results from VideoMaMa, we require input masks for all frames in the video sequence. We employ a point prompt on the first frame, which is then propagated throughout the entire video using SAM2's tracking capability.

For quantitative evaluation on benchmark datasets, we binarize the ground-truth mask from the first frame to serve as the initial prompt for SAM2, subsequently propagating it across all remaining frames to generate the input mask sequence for VideoMaMa.

In the VideoMaMa architecture, we extract semantic features using DINOv2~\cite{dinov3} and inject them into the first upsampling block (Upblock 1) of the Stable Video Diffusion (SVD)~\cite{svd} backbone. The feature injection is performed through a two-layer MLP projection module that aligns the DINOv3 feature dimensions with the SVD decoder architecture.

\paragraph{SAM2-Matte}
We implement SAM2-Matte based on the SAM2~\cite{sam2} Base-Plus architecture, initialized with official pretrained weights. The model is trained for $100{,}000$ iterations with a batch size of $4$, employing a mixed training strategy that combines both video and image samples. The composition of existing datasets utilized in this training phase is detailed in Table~\ref{tab:existing_datasets}. For optimization, we adopt a combination of L1 and Laplacian pyramid losses. The base learning rate is set to $5\times10^{-6}$, while a separate learning rate of $3\times10^{-6}$ is applied specifically to the image encoder. Training is conducted with a cosine learning rate scheduler.

\begin{table}[t!]
\centering
\caption{\textbf{Summary of the datasets} constituting the ``Existing Dataset'' used to train SAM2-Matte.}
\label{tab:existing_datasets}
\resizebox{\columnwidth}{!}{%
  \begin{tabular}{@{}ll@{}}
  \toprule
  \textbf{Category} & \textbf{Datasets} \\
  \midrule
  Image       & DVM~\cite{DVM}, AM-2k~\cite{am2k}, DAViD~\cite{DAViD}, P3M-10k~\cite{p3m}, RefMatte~\cite{refmatte} \\
  Video       & CRGNN~\cite{crgnn}, VideoMatte240K~\cite{video240k}, DVM~\cite{DVM} \\
  Background & VideoMatting108~\cite{video108}, DVM~\cite{DVM} \\
  \bottomrule
  \end{tabular}%
}
\end{table}

\section{Evaluation Metric Details}
\label{sec:MAD-T}
Evaluating matte accuracy in transition regions is critical for video matting, particularly for capturing fine details like hair or motion blur. Standard metrics often dilute these errors by averaging over large background regions. To address this, we evaluate MAD-T, which restricts the Mean Absolute Difference calculation specifically to the unknown region. The complete calculation procedure is detailed in Algorithm~\ref{alg:mad_t_video}. As ground-truth trimaps are rarely available for video datasets, we generate pseudo-trimaps dynamically for each frame $t$. We derive binary foreground and background masks from the ground-truth alpha $\alpha_{gt}^{(t)}$ and apply morphological erosion using a $10 \times 10$ elliptical structuring element. The unknown region $\mathcal{U}^{(t)}$ is defined as the set of pixels excluded from both eroded masks. The final MAD-T score is reported as the temporal average of the per-frame error within $\mathcal{U}^{(t)}$, scaled by $10^3$.

\begin{algorithm}[t]
\caption{Frame-wise Calculation of MAD-T}
\label{alg:mad_t_video}
\begin{algorithmic}[1]
\Require Pred. $\hat{\alpha}_t$, GT $\alpha_{gt,t}$, Kernel $k=10$
\Ensure MAD-T Score $\mathcal{S}_t$ for frame $t$

\State \textbf{Initialize} structuring element $K \leftarrow \text{Ellipse}(k, k)$

\State \Comment{Define Certain Regions via Erosion}
\State $\mathcal{M}_{fg} \leftarrow \mathbb{I}(\alpha_{gt,t} = 255)$ \hfill \Comment{Binary Foreground}
\State $\mathcal{M}_{bg} \leftarrow \mathbb{I}(\alpha_{gt,t} = 0)$ \hfill \Comment{Binary Background}
\State $\mathcal{C}_{fg} \leftarrow \mathcal{M}_{fg} \ominus K$ \hfill \Comment{Erode to ensure certainty}
\State $\mathcal{C}_{bg} \leftarrow \mathcal{M}_{bg} \ominus K$ \hfill \Comment{Erode to ensure certainty}

\State \Comment{Isolate Unknown Region (Trimap)}
\State $\mathcal{U}_t \leftarrow \neg (\mathcal{C}_{fg} \cup \mathcal{C}_{bg})$
\State $N_t \leftarrow \sum \mathcal{U}_t$ \hfill \Comment{Pixel count in unknown region}

\State \Comment{Compute Error in Unknown Region}
\If{$N_t > 0$}
    \State $\mathcal{D}_t \leftarrow |\hat{\alpha}_t - \alpha_{gt,t}|$
    \State $\mathcal{S}_t \leftarrow \frac{1}{N_t} \sum_{(x,y) \in \mathcal{U}_t} \mathcal{D}_{t,x,y} \times 1000$
\Else
    \State $\mathcal{S}_t \leftarrow 0$
\EndIf

\State \Return $\mathcal{S}_t$
\end{algorithmic}
\end{algorithm}

\section{Comparison with SAM2}
\label{sec:sam2_comparison}
\begin{figure}
    \centering
    \includegraphics[width=1\columnwidth]{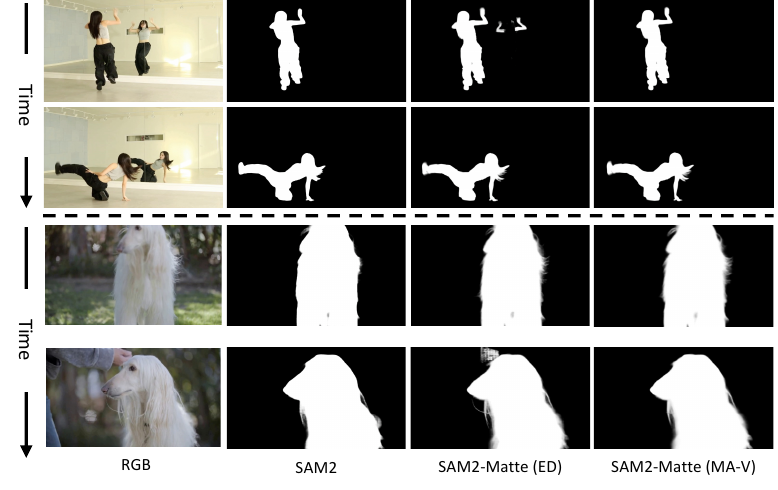}
    \caption{\textbf{Comparison on a real video between SAM2-Matte and SAM2~}\cite{sam2} with sigmoid function on mask logit to simulate alpha matte generation. ED denotes that existing dataset.}
    \label{fig:sam_comparison}
\end{figure}
SAM2~\cite{sam2} generates binary masks by computing the dot product between image features and prompt tokens that undergo cross-attention in SAM's decoder blocks, followed by thresholding at a specified value. By removing the thresholding operation and retaining only the sigmoid activation, the model can simulate alpha matte generation by mapping the mask logits to the continuous range [0,1], which forms the basis of our SAM2-Matte approach. We compare our SAM2-Matte with the original SAM2 in Figure~\ref{fig:sam_comparison} to demonstrate the effectiveness of the \mbox{MA-V} dataset, illustrating that without fine-tuning on \mbox{MA-V}, SAM2 cannot generate robust video matting results.

\section{Limitation}
\label{sec:limitation}
\begin{figure}[t!]
    \centering
    \includegraphics[width=1.0\columnwidth]{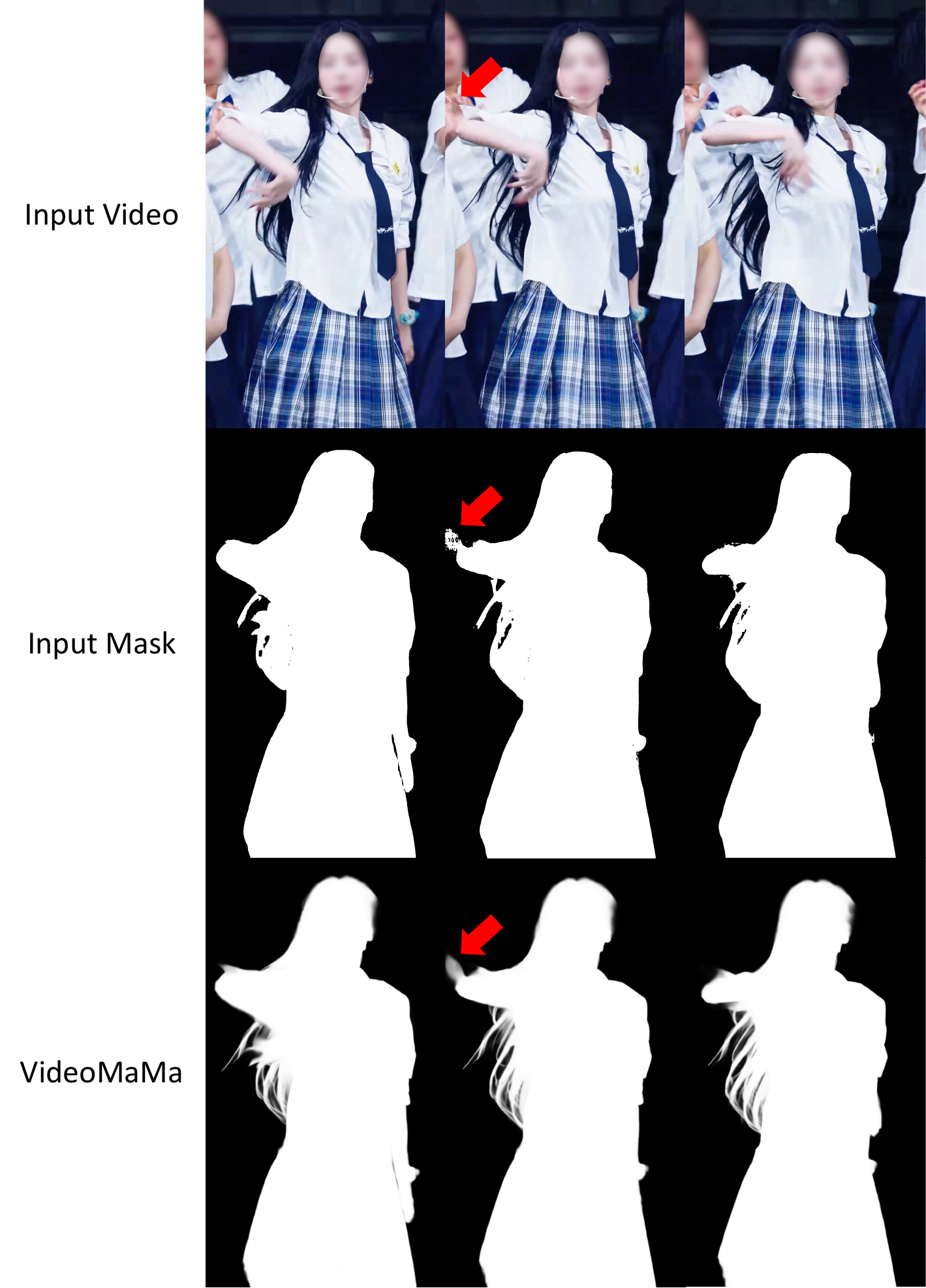}
    \caption{\textbf{Limitations.} As highlighted by the red arrows, the model struggles to refine the matte when the input guidance mask exhibits significant errors from the input mask.}
    \label{fig:limitation}
    \vspace{-10pt}
\end{figure}
\paragraph{VideoMaMa}
While VideoMaMa can generate high-quality video mattes from rough or imperfect masks, it cannot generate mattes for regions where completely wrong mask is provided (\emph{e.g.} capturing wrong instance). As shown in Figure~\ref{fig:limitation}, when the input mask incorrectly captures the wrong instance, VideoMaMa propagates this error to the output. Our method successfully refines masks that lack fine-grained details but struggles when the input mask is fundamentally incorrect, such as when it captures an entirely different object instance. This limitation is inherent to mask-guided approaches, as the model relies on the mask to define the target foreground object.

\paragraph{SAM2-Matte}
Conducting alpha matte generation in high-resolution is essential for getting high quality alpha matte. However, the standard SAM2 architecture has a structural limitation for this task due to its mask decoder design. Specifically, SAM2 generates segmentation masks at a resolution of $64\times64$, which is subsequently upsampled to match the input dimensions. This resolution is substantially lower than those employed by other matting methods~\cite{zim, matanyone}, which typically predict alpha mattes at significantly higher resolutions to preserve fine-grained details such as hair strands and object boundaries. Consequently, directly applying SAM2 for video matting results in the loss of critical high-frequency information necessary for accurate alpha matte estimation. This architectural constraint poses a fundamental challenge for adapting SAM2 to the video matting task, where precise boundary delineation and detail preservation are paramount in getting alpha value of [0,1].

\newpage
{
    \small
    \bibliographystyle{ieeenat_fullname}
    \bibliography{main}
}


\end{document}